\documentclass[pmlr]{jmlr}% new name PMLR (Proceedings of Machine Learning)

\RequirePackage{graphicx}
 \usepackage{booktabs}
 \usepackage{longtable}
 \usepackage{multirow}
 \usepackage{float}
 \usepackage[normalem]{ulem}

 \usepackage{xcolor}
 \usepackage{enumitem}

\makeatletter
\def\set@curr@file#1{\def\@curr@file{#1}} %temp workaround for 2019 latex release
\makeatother
\usepackage[load-configurations=version-1]{siunitx} % newer version

\newcommand{\response}[1]{#1}

\theorembodyfont{\upshape}
\theoremheaderfont{\scshape}
\theorempostheader{:}
\theoremsep{\newline}

\jmlrproceedings{}{}
\jmlrvolume{}
\jmlrpages{}
\jmlryear{2026}
\jmlrworkshop{}

\title[Imitation Learning for Robot Assistance in Open Surgery]{Imitation Learning for Robot Assistance in Open Surgery: A Multi-Policy Evaluation on Suture Following}

\author[Wang, Yang, et al.]{%
  \Name[X.~Wang]{Xucheng Wang, BS\textsuperscript{*,1}} \Email{davidx\_wang@hms.harvard.edu}\\
  \Name[Z.~Yang]{Zhizhou Yang, MD\textsuperscript{*,1,2}} \Email{zyang17@mgh.harvard.edu}\\
  \Name[X.~Zhang]{Xiaoman Zhang, PhD\textsuperscript{1}} \Email{xiaoman\_zhang@hms.harvard.edu}\\
  \Name[S.E.~Kim]{Sung Eun Kim, MD\textsuperscript{1}} \Email{sungeun\_kim2@hms.harvard.edu}\\
  \Name[R.~Hardy]{Romain Hardy, MS\textsuperscript{1}} \Email{romain\_hardy@g.harvard.edu}\\
  \Name[P.~Rajpurkar]{Pranav Rajpurkar, PhD\textsuperscript{1}}\\
  \addr \textsuperscript{1}Department of Biomedical Informatics, Harvard Medical School, Boston, MA\\
  \addr \textsuperscript{2}Department of Surgery, Massachusetts General Hospital, Boston, MA\\
  \addr \textsuperscript{*}These authors contributed equally to this work.}

\begin{document}

\maketitle

\begin{abstract}
% Learning-based surgical automation has focused almost exclusively on minimally invasive platforms and on increasing levels of robot autonomy as primary surgeon. Open surgery assistance, where currently at least one human performs critical ancillary tasks alongside the primary surgeon, remains unexplored in the realm of surgical automation. 
% Learning-based surgical automation has focused almost exclusively on minimally invasive platforms and on increasing the robot's autonomy as the primary operator. Open surgery assistance, where a human assistant performs critical tasks alongside the primary surgeon, remains unexplored. 
This study presents the first evaluation of general-purpose \textbf{imitation learning for \emph{surgeon--robot collaborative} assistance in open surgery}, targeting \emph{suture following}: the grab-pull-release motion an assistant performs at every stitch. 
We collect 160 teleoperated demonstrations (32{,}374 frames) on an open-source robot arm, benchmark four architecturally diverse imitation learning policies (ACT, Diffusion Policy, SmolVLA, $\pi_0$) across 28 trained models evaluated in 32 configurations along three clinically motivated dimensions: \emph{dataset size}, \emph{camera viewpoint}, and \emph{background variation}.
Our results demonstrate that under ideal conditions \response{on a controlled benchmark}, the four policies achieve 50--\textbf{75\%} task success, with depth error as the dominant failure mode across all architectures. 
Among all policies, $\pi_0$ achieves the strongest results with a pretrained vision–language backbone, demonstrating superior data efficiency, greater robustness to background variation, and smoother trajectories compatible with surgical workflow. 
When deployed in a \response{live in-loop} surgeon–robot suturing trial, $\pi_0$ yields a \textbf{92\% stitch completion rate}. 
These findings establish collaborative robotic assistance in open surgery as a feasible target for imitation learning and highlight depth perception and end-effector design as key priorities for clinical translation. All code, data, and model weights are released at \url{https://github.com/rajpurkarlab/Suture_Follow_Evaluation}.

% which leverages a pretrained vision–language backbone, achieves the strongest results: it yields superior data efficiency, retains 50\% success under unseen backgrounds where policies trained from scratch collapse to 10\%, and produces smoother trajectories more compatible with live surgical workflow. 
% $\pi_0$ fine-tuned on our demonstrations reaches 75\% success on a controlled benchmark, retains 50\% success under unseen backgrounds showing that pretrained vision--language backbones are particularly valuable for data efficiency and visual distribution shift. 
\end{abstract}

%% ====================================================================
\section{Introduction}
%% ====================================================================

\noindent \textbf{Collaborative Open Surgery.} Open surgery has been the primary treatment for the majority of surgical conditions worldwide \citep{schneider2021inequalities, mattingly2022mis}, and is inherently collaborative: while the primary surgeon focuses on high-stakes dissection or reconstruction, an assistant provides exposure and performs critical assisting maneuvers at every step \citep{catchpole2008teamwork, gillespie2013team}.
This collaboration is especially evident during suturing: in a running closure or anastomosis, the assistant follows the suture, maintaining gentle tension and organizing the thread at each stitch so the surgeon can work efficiently.
% While minimally invasive surgery (MIS) has expanded rapidly for eligible intra-abdominal and intra-thoracic procedures, open operations remain necessary and represent in fact the vast majority of all surgical cases given anatomical and exposure need. 
% Open surgery is inherently collaborative: while the primary surgeon focuses on high-stakes dissection or reconstruction, an assistant provides exposure, maintains tension, and performs critical assisting maneuvers at every step \citep{catchpole2008teamwork, gillespie2013team}. 
% Suturing illustrates this well. During a running closure or anastomosis, the assistant follows the suture, maintaining gentle tension and organizing the thread at each stitch so the surgeon can work efficiently.
This grab-pull-release cycle is essential, highly repetitive, and a natural target for automation, particularly given the global shortage of trained surgical personnel \citep{meara2015global, holmer2015global, alkire2015global, kewalramani2025workforce} and the limited availability of skilled assistants, even in well-resourced settings.

\begin{figure}[!htbp]
  \centering
  \includegraphics[width=0.95\textwidth]{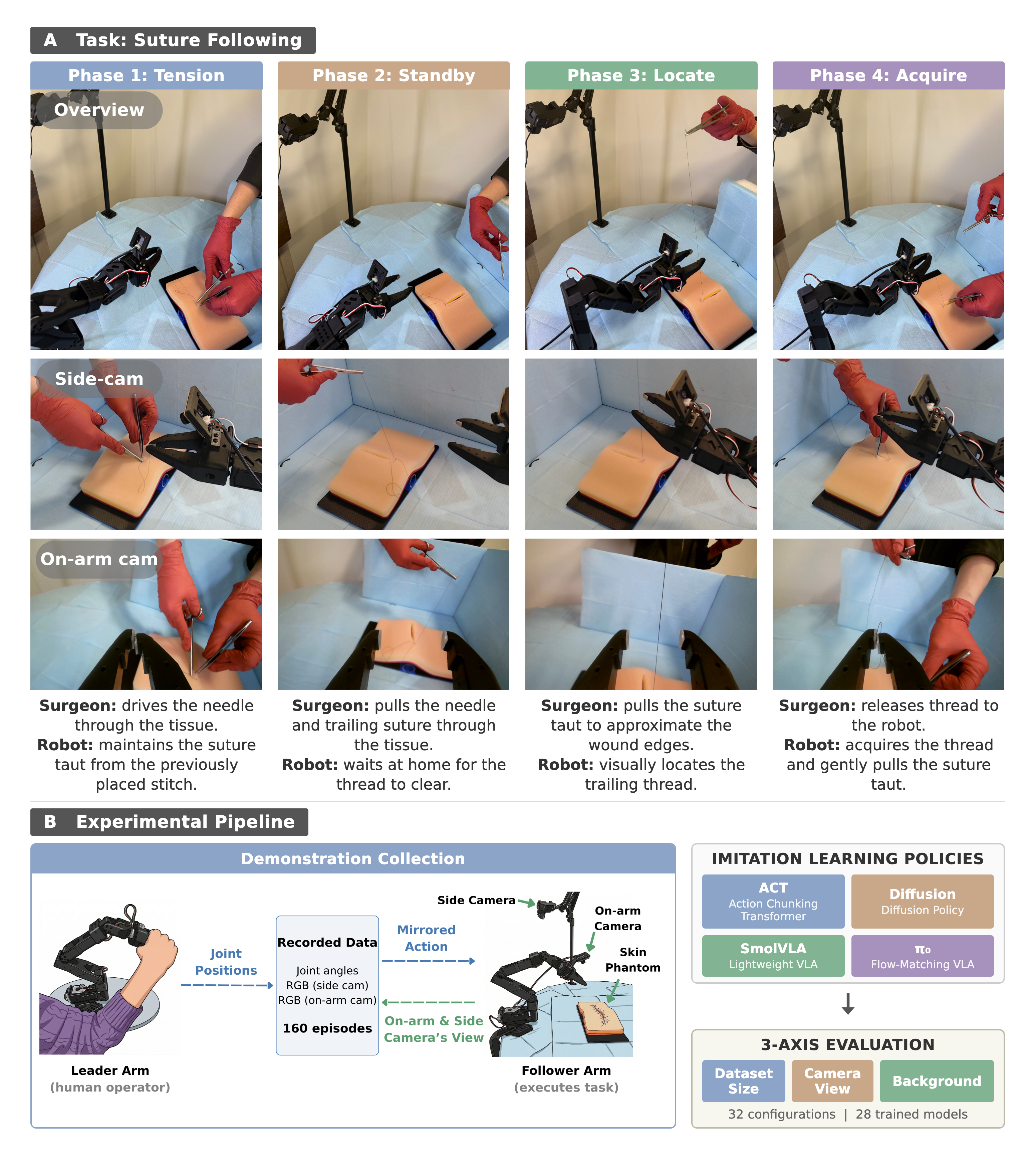}
  \caption{Overview of the suture following task and experimental pipeline. \textbf{(A)} Four phases of a single suture-following cycle, shown from three camera views (overview, side, on-arm): Phase~1, the robot holds the suture taut while the surgeon drives the needle; Phase~2, the surgeon pulls the suture through as the robot returns home; Phase~3, the surgeon tightens the stitch while the robot locates the thread; Phase~4, the surgeon releases the thread to the robot, which grasps and tensions it before the cycle repeats. 
  % Four phases of a single suture-following cycle, shown from three camera views (overview, side, on-arm). The cycle repeats for every stitch in a running closure: Phase~1, the robot maintains the suture taut while the surgeon drives the needle through tissue; Phase~2, the surgeon pulls the suture through and the robot returns to its home position; Phase~3, the surgeon pulls the suture taut to approximate the wound edges while the robot visually locates the thread; Phase~4, the surgeon releases the thread to the robot, which acquires it and gently pulls the suture taut, then the cycle returns to Phase~1. 
  \textbf{(B)} 160 teleoperated demonstrations collected on open-source hardware are used to train four imitation learning policies, which are then evaluated across three axes in 32 configurations.}
  \label{fig:overview}
\end{figure}

% Existing robotic surgical systems do not address this need. 
% The dominant paradigm, exemplified by the da Vinci platform, enhances the primary surgeon's capabilities within the \textbf{minimally invasive surgery (MIS)} setting \citep{sheetz2020trends, barbash2010technology}. 
% Learned surgical automation has advanced rapidly on MIS platforms \citep{shademan2016supervised, srt2024, kim2025srth}, but these systems depend on the structured video and kinematic data that such platforms provide and are not transferable to open surgical workflows \citep{schmidgall2025surgicalreview}. Moreover, much of the literature \citep{schmidgall2025surgicalreview, khanna2025cuttingedge} pursues increasing levels of robot autonomy, with the long-term goal of the robot performing procedures independently. 
% A limited body of recent work has begun to explore robotic surgical assistance \citep{robonurse2024, moeact2026}, but these efforts do not address intra-procedural assistant roles in \textbf{open surgery}.

% \xiaoman{I feel it is better to put the discussion of MIS in the Related Work section. In intro, probably only a sentence for comparasion with our setting. }
\paragraph{Gaps in Existing Robotic Systems.}
Existing robotic surgical systems have largely focused on a different clinical setting: either enhancing or automating the primary surgeon's tasks within the {minimally invasive surgery (MIS)} paradigm \citep{sheetz2020trends, barbash2010technology, schmidgall2025surgicalreview, khanna2025cuttingedge}. Intra-procedural assistant roles in {open surgery}, by contrast, remain largely unstudied. In this work, we explore \textbf{\emph{surgeon--robot collaboration} for open surgery}. The robot takes on the surgical assistant's role, performing the repetitive supporting maneuvers at the bedside, while the surgeon retains the primary operative tasks, such as driving the needle and placing stitches. This keeps the robot's responsibilities well-defined and lower-risk, while addressing a surgical setting that existing robotic platforms do not serve.

\paragraph{Imitation Learning for Surgical Assistance.}
\response{Realizing this collaboration requires a robot that can learn assistive behaviors from demonstration, and recent advances in imitation learning have demonstrated their feasibility:} open-source hardware and learning frameworks now enable a robot to acquire dexterous manipulation skills by observing human demonstrations \citep{zhao2023learning, fu2024mobile, cadene2024lerobot}, and powerful policy architectures \citep{chi2023diffusion, lee2024behavior, black2024pi0, smolvla2025} have pushed these capabilities to increasingly complex tasks.
However, these approaches have seen limited application in surgical assistance: existing benchmarks \citep{ahmidi2017dataset, xu2021surrol, yu2024orbit} all target the surgeon's role on MIS platforms, and no head-to-head comparison of general-purpose policies exists for any surgical assistive task.

% Our work takes a different approach: rather than automating the surgeon, we study \emph{surgeon--robot collaboration} for open surgery, in which the surgeon retains full control of all primary tasks while a robot handles the assistant's repetitive role alongside them. 
% A small body of work has begun to explore robotic surgical assistance \citep{robonurse2024, moeact2026}, but these efforts remain within the MIS paradigm; collaborative assistance in open surgery, and suture following in particular, has not been studied.

% In parallel, the general robotics community has developed open-source hardware and \emph{imitation learning} frameworks, in which a robot learns to perform a task by observing human demonstrations rather than through explicit programming \citep{zhao2023learning, fu2024mobile, cadene2024lerobot}. Combined with powerful policy architectures \citep{chi2023diffusion, lee2024behavior, black2024pi0, smolvla2025}, these methods now solve increasingly dexterous manipulation tasks. This capability has not been applied to surgical problems: existing benchmarks \citep{ahmidi2017dataset, xu2021surrol, yu2024orbit} all target the surgeon's role on MIS platforms, and no head-to-head comparison of general-purpose policies exists for any surgical assistive task.

% In this work, we bring these capabilities to bear on open surgical assistance for the first time (Figure~\ref{fig:overview}).
\paragraph{Our Approach and Contributions.}
In this work, we introduce suture following as a clinically grounded case study for surgeon--robot collaboration. Using an accessible, non-specialized open-source robotic platform, we collect 160 teleoperated demonstrations and conduct a systematic evaluation of four architecturally diverse imitation learning policies (ACT \citep{zhao2023learning}, Diffusion Policy \citep{chi2023diffusion}, SmolVLA \citep{smolvla2025}, and $\pi_0$ \citep{black2024pi0}) across three clinically motivated axes: \emph{dataset size}, \emph{camera viewpoint}, and \emph{background variation} (Figure~\ref{fig:overview}). This offers an initial empirical comparison of how general-purpose imitation learning frameworks transfer to the precision requirements of surgical assistance, characterizing each architecture's distinct failure modes and data efficiency. \response{The full demonstration dataset, training and evaluation code, and trained model weights are available at \url{https://github.com/rajpurkarlab/Suture_Follow_Evaluation}.}

\subsection*{Generalizable Insights about Machine Learning in the Context of Healthcare}

\begin{itemize}[leftmargin=*,itemsep=2pt]
    \item \textbf{Imitation learning for intra-operative surgeon--robot collaboration in open surgery:} We show the potential of imitation learning to support a robot taking on the surgical assistant's role during an active open surgical procedure, establishing collaborative open-surgical assistance as a feasible and clinically relevant application domain for the ML-for-health community.
    \item \textbf{Pretrained vision--language backbones for clinical robustness:} Our evaluation along three clinically motivated axes reveals that policies with pretrained vision--language backbones require fewer demonstrations to reach competent performance and degrade more gracefully under visual distribution shifts (camera viewpoint, background variation). This suggests that foundation-model-based architectures are particularly well suited for surgical environments, where visual conditions at deployment inevitably differ from those seen during data collection.
\end{itemize}

%% ====================================================================

\section{Related Work}
%% ====================================================================

\paragraph{Surgical autonomy on MIS platforms.}
The dominant robotic surgical paradigm, exemplified by the da Vinci platform \citep{sheetz2020trends, barbash2010technology}, is designed around the primary surgeon operating within minimally invasive surgery. Building on this hardware, learned surgical automation has progressed significantly on the da Vinci Research Kit (dVRK), from STAR's supervised soft-tissue suturing \citep{shademan2016supervised} to SRT/SRT-H's multi-task autonomy via imitation learning \citep{srt2024, kim2025srth} and SutureBot's end-to-end suturing benchmark \citep{suturebot2025}. However, these methods rely on MIS consoles, limiting their direct transfer to open surgical workflows \citep{schmidgall2025surgicalreview}. The literature targets the \emph{surgeon's} tasks (needle driving, knot tying) and pursues increasing levels of autonomous operation \citep{khanna2025cuttingedge}; the \emph{assistant's} subtasks, which are more repetitive and lower-risk, have received far less attention.

\paragraph{Robotic surgical assistance.}
A nascent line of work has begun to target assistance rather than full autonomy. RoboNurse-VLA \citep{robonurse2024} automates the scrub nurse's instrument handover role, which is independent from the surgical assistant and does not involve intra-procedural coordination with the surgeon's actions on tissue. MoE-ACT addresses tissue retraction on the dVRK \citep{moeact2026} and \citet{long2025surgicalembodied} demonstrates zero-shot sim-to-real for laparoscopic assistive tasks, but both operate within MIS through a teleoperated console rather than at the bedside. Collaborative intra-operative assistance in open surgery remains unstudied.

\paragraph{Imitation learning policies.}
We evaluate four policies spanning three architectural families: ACT \citep{zhao2023learning}, a conditional-VAE transformer; Diffusion Policy \citep{chi2023diffusion}, which models trajectories via conditional denoising diffusion; $\pi_0$ \citep{black2024pi0}, a large-scale vision-language-action (VLA) flow model pretrained on diverse multi-robot data; and SmolVLA \citep{smolvla2025}, a lightweight open-source VLA. DP3 \citep{ze2024dp3} extends diffusion-based policies to 3D point clouds. No prior work has systematically compared these architectures on a surgical task.

\paragraph{Surgical benchmarks on accessible robot platforms.}
Existing surgical benchmarks (JIGSAWS \citep{ahmidi2017dataset}, SurRoL \citep{xu2021surrol}, ORBIT-Surgical \citep{yu2024orbit}, SurgicAI \citep{schmidgall2024surgicai}) are simulation-only or dVRK-specific. The LeRobot framework \citep{cadene2024lerobot} and ALOHA/Mobile ALOHA hardware \citep{zhao2023learning, fu2024mobile} have established that open-source platforms support learned dexterous manipulation in general robotics; our work applies this ecosystem to open surgical assistance for the first time.

\section{Methods}

\subsection{Task Definition: Suture Following}
\label{sec:task_def}

During a running suture closure or anastomosis, the primary surgeon's two hands are constantly occupied: one holds the needle driver to drive the curved needle through the tissue, and the other holds the forceps to grab and stabilize the target tissue, followed by needle retrieval and reload. Between every stitch, the trailing thread oftentimes needs to be pulled aside and held under appropriate tension. Without this, the previous stitch slackens, the thread tangles or drifts back into the operating field, resulting in unintended locking, and the closure cannot be tightened before the next bite. 
In current practice, this is the job of a human assistant, who tracks the moving suture, grasps it near the trailing end, and pulls it out of the way until the surgeon is ready to drive the next stitch. 
We define the \textbf{suture following} task as the robotic counterpart of this assistive role: given a suture thread trailing from a sutured section of a silicone tissue phantom, the robot must visually locate the thread, grasp it, and pull it taut. 
The suture thread is thin and low-contrast, and its position varies between trials, so successful acquisition requires both sub-centimeter precision and reliable depth reasoning.

\subsection{Hardware Platform}

We use a low-cost robotic system built around an SO-101 follower arm \citep{cadene2024lerobot}, an open-source design from the LeRobot hardware ecosystem with five actuated revolute joints and a 1-DoF parallel-jaw gripper driven by six servomotors. Visual observations come from two commodity USB webcams, a wrist-mounted (on-arm) camera and a fixed side-view camera, each streaming RGB at 10\,Hz. Demonstrations are collected via teleoperation: a human operator physically moves a second SO-101 arm (the \emph{leader}), and the task-performing arm (the \emph{follower}) mirrors its joint positions in real time, recording the resulting motions as training data (Figure~\ref{fig:overview}B). 
% The entire system is low-cost, accessible, and fully open-source.

\subsection{Task Environment}

The task environment, illustrated in Figure~\ref{fig:overview}A, consists of a silicone skin pad serving as the tissue phantom, into which a standard surgical silk 3/0 suture has been placed in a running configuration so that a length of thread trails from the most recent stitch and is available for the robot to acquire. 
All training demonstrations are collected in a \emph{clean background} condition, in which the workspace surrounding the phantom is covered by a standard medical blue surgical drape. To probe environmental robustness at evaluation time, we additionally test the trained policies in a \emph{randomized background} condition, in which the blue surgical drape is removed to expose the underlying room environment (tabletop, instruments, and other lab clutter) around the phantom.
% The randomized condition is used only at evaluation; no policy is exposed to it during training.

\subsection{Data Collection}

We collected 160 teleoperated episodes by a two-person team: a single teleoperator drove the leader arm throughout the entire dataset to keep motion style consistent, while a second person handled the surgical-side tasks (driving the needle through the phantom and presenting the trailing thread). Across episodes, we deliberately varied the suture direction and the position of the phantom on the OR table so that the dataset covers a range of starting configurations rather than a single canonical setup. All demonstrations were collected in the clean background condition.
The resulting dataset (Figure~\ref{fig:overview}B) contains 32{,}374 RGB-state frames in total, recorded at 10\,Hz, corresponding to under one hour of pure demonstration time. 
Each frame stores synchronized camera observations and the 6-dimensional joint state (5 arm joints + gripper); actions are the corresponding 6-dimensional leader-arm joint positions at the same cadence. No data augmentation or filtering is applied beyond the LeRobot dataset normalization (per-feature mean/std and min/max statistics computed from the training split).

\subsection{Imitation Learning Policies}

\response{We select four state-of-the-art, architecturally diverse policies as representatives of the main imitation-learning families, for broad coverage of current policy designs rather than practical convenience:} ACT \citep{zhao2023learning}, Diffusion Policy \citep{chi2023diffusion}, $\pi_0$ \citep{black2024pi0}, and SmolVLA \citep{smolvla2025}, spanning transformer-based, diffusion-based, and vision-language-action architectures:

\begin{itemize}[leftmargin=*,itemsep=4pt]
    \item \textbf{ACT (Action Chunking with Transformers).} A conditional VAE with a transformer backbone that predicts temporally extended action chunks via a single forward pass. We use the LeRobot default configuration with a ResNet-18 visual encoder ($\sim$52\,M parameters), trained from scratch on our dataset.

    \item \textbf{Diffusion Policy.} A visuomotor policy that models action trajectories as a conditional denoising diffusion process, using a CNN-based visual encoder ($\sim$263\,M parameters). We train the LeRobot default configuration from scratch on our dataset.

    \item \textbf{$\pi_0$.} A vision-language-action flow-matching model ($\sim$3.5\,B parameters) pretrained on a large corpus of diverse multi-robot manipulation data. We fine-tune $\pi_0$ on our dataset using the LeRobot default configuration, keeping the PaliGemma vision-language backbone frozen and updating only the action expert head.

    \item \textbf{SmolVLA.} A lightweight open-source vision-language-action model ($\sim$450\,M parameters) designed for accessible robotics research. We fine-tune SmolVLA on our dataset using the LeRobot default configuration, keeping the vision-language backbone frozen and updating only the action expert.
\end{itemize}

\noindent All four policies share the same RGB observation inputs from the on-arm and side cameras, the same action space (joint positions and gripper state), and are trained with the LeRobot default recipe for each architecture on a single NVIDIA GPU.
Full hyperparameters and training time per model are available in the supplementary material.

\subsection{Experimental Design}
\label{sec:experimental_design}

We evaluate along three orthogonal axes, each probing a different practical concern for clinical deployment:

\begin{itemize}[leftmargin=*,itemsep=3pt]
    \item \textbf{Dataset size.} Expert surgical demonstrations are expensive to collect. We train each of the four policies on \{20\%, 40\%, 60\%, 80\%, 100\%\} of the full 160-episode dataset (i.e., 32, 64, 96, 128, and 160 episodes) to characterize data efficiency curves and identify minimum viable dataset sizes.

    \item \textbf{Camera viewpoint configuration.} Operating rooms impose constraints on camera placement. We compare three configurations: \textbf{dual camera} (on-arm + side), the full observation setup; \textbf{on-arm only}, simulating a wrist-mounted endoscopic-style view; and \textbf{side camera only}, simulating a fixed external camera. All viewpoint ablations use 100\% of the training data to isolate the effect of visual input.

    \item \textbf{Environmental robustness.} To probe generalization, we test baseline models (dual camera, 100\% data) in a \textbf{randomized background} condition, where the workspace background and surrounding objects are varied between test episodes. This simulates the visual diversity encountered in real clinical settings.
\end{itemize}

\noindent In total, we train \textbf{28 models}: $5 \times 4 = 20$ (dataset-size ablations across four policies) $+ 2 \times 4 = 8$ (single-camera variants for each policy at 100\% data).

\subsection{Evaluation Protocol}
\label{sec:eval_protocol}

We evaluate all models under a \emph{controlled benchmark} protocol designed to isolate policy performance from surgeon variability. Each model is evaluated over 20 test episodes. In each episode, a human tester holds the trailing suture thread taut and stationary in a randomized direction, and the tissue phantom is placed at a random position within the workspace. The robot executes autonomously: it must visually locate the thread, navigate to it, and acquire it in the gripper without any intervention. This controlled setting provides a reproducible, strict measure of each policy's visuomotor competence and is complemented by the in-loop surgeon--robot trial described in Section~\ref{sec:inloop}. \response{As this is the first study of automated suture following, no prior human or robotic performance baseline exists for direct comparison; the controlled protocol therefore serves as a reference point for future work.}

\paragraph{Success criteria.} An episode is considered successful if the robot successfully acquires the suture thread in the gripper, completing the visually guided reach-and-grasp motion described in Section~\ref{sec:task_def}. 

\paragraph{Failure mode taxonomy.} To provide diagnostic insight beyond binary success/failure, we categorize each failed episode into one of three failure modes by visual inspection:

\begin{itemize}[leftmargin=*,itemsep=3pt]
    \item \textbf{Lateral error.} The robot approaches with a significant horizontal offset (too far left or right), missing the thread. This indicates a failure in visual localization of the thread in the image plane.
    \item \textbf{Depth error.} The robot does not reach far enough (undershoot) or extends too far past the thread (overshoot) along the approach axis. In both cases, the gripper fails to acquire the thread at the correct depth. This suggests difficulty in depth estimation for thin, low-contrast objects in 3D space.
    \item \textbf{Task breakdown.} The robot fails to produce a coherent attempt at the task. We classify an episode as a task breakdown if any of the following holds: (i) the arm does not initiate motion toward the thread; (ii) the arm drifts to an unrelated region of the workspace; or (iii) the arm enters a repetitive loop of motions without progress toward the thread. Unlike lateral and depth errors, which reflect localization inaccuracies in an otherwise reasonable approach trajectory, task breakdowns indicate a fundamental failure of the policy to map the current observation to a meaningful action sequence.
\end{itemize}

%% ====================================================================
\section{Results}
\label{sec:results}
%% ====================================================================

% \begin{table}[t]
%   \centering
%   \small
%   \caption{Baseline performance: dual camera, 100\% training data, clean background. Failure modes are reported as a percentage of all test episodes. \xiaoman{Table 1 is redundant? all numbers are covered by figure2 and figure3.}}
%   \begin{tabular*}{\textwidth}{@{\extracolsep{\fill}}lcccc}
%   \toprule
%     \textbf{Policy} & \textbf{Success (\%)} & \textbf{Lateral (\%)} & \textbf{Depth (\%)} & \textbf{Breakdown (\%)} \\
%     \midrule
%     ACT              & 50 & 25 & 25 & 0 \\
%     Diffusion Policy & 65 &  0 & \textbf{35} & 0 \\
%     SmolVLA          & 65 & 10 & 25 & 0 \\
%     $\pi_0$          & \textbf{75} &  5 & 20 & 0 \\
%     \bottomrule
%   \end{tabular*}
%   \label{tab:baseline}
% \end{table}

\subsection{Baseline Performance}

\begin{figure}[!htbp]
  \centering
  \includegraphics[width=0.9\textwidth]{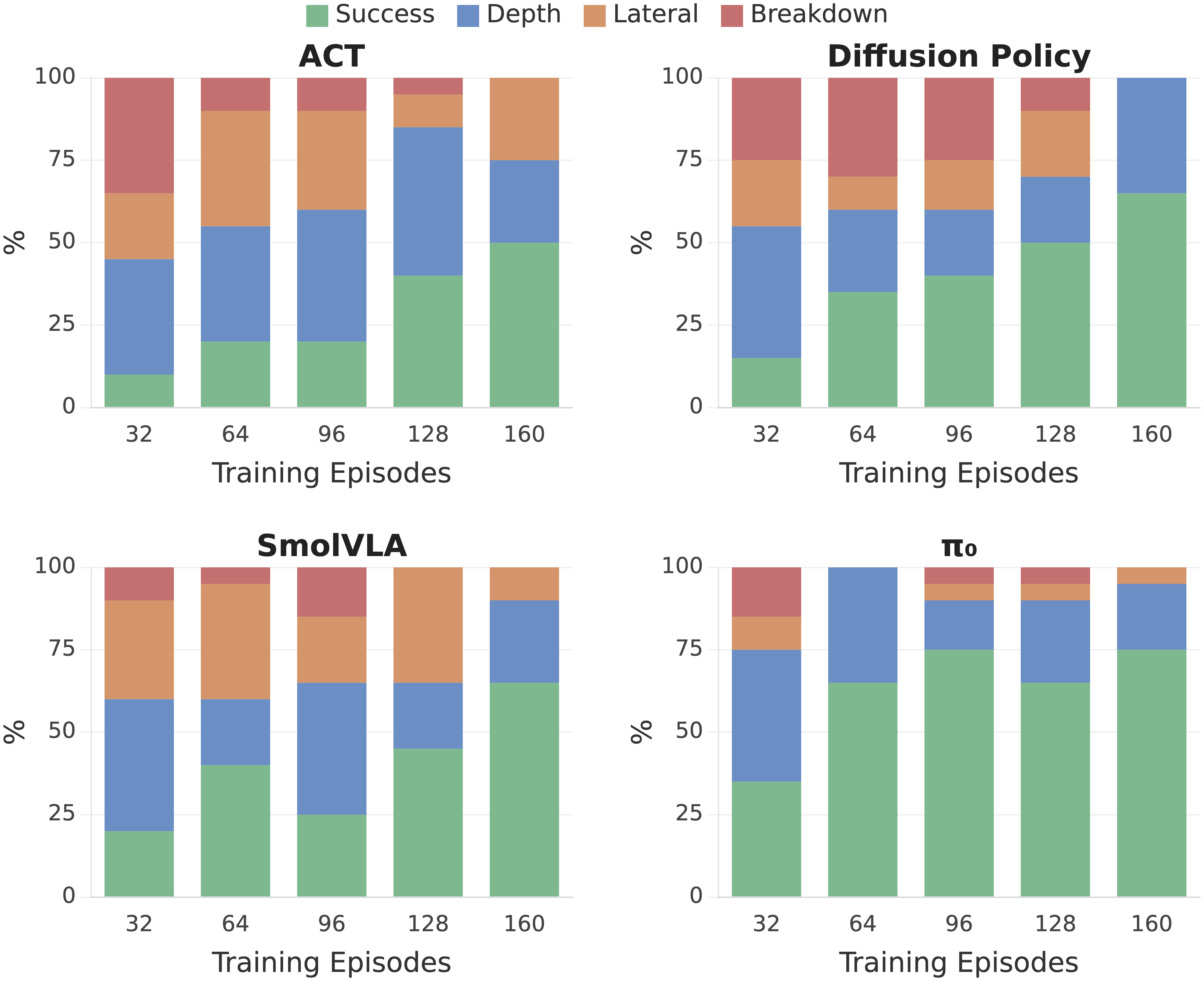}
  \caption{Failure mode distribution across training dataset sizes for each policy. Green indicates successful episodes; blue, orange, and red indicate depth error, lateral error, and task breakdown, respectively. At low data regimes, task breakdowns (red) are prevalent across all policies, while at higher data regimes failures shift toward depth and lateral errors.}
  \label{fig:failure_modes}
\end{figure}

Figure~\ref{fig:failure_modes} contains the baseline performance of all four policies under ideal conditions (dual camera, 100\% training data: 160 Episodes, clean background).
Under ideal conditions, the four policies form a clear ordering: ACT performs the worst at 50\% task success, Diffusion Policy and SmolVLA tie in the middle at 65\%, and $\pi_0$ achieves the highest success rate at 75\%. Notably, the smaller SmolVLA only matches Diffusion Policy, while the much larger $\pi_0$ pulls ahead by another 10 points, suggesting that both scale and pretraining of the vision-language backbone contribute to baseline competence on this task. \response{This ranking is specific to our precision-critical, low-data setting: on general-purpose multi-task simulation benchmarks (e.g., LIBERO \citep{liu2023libero}, Meta-World \citep{yu2020metaworld}), SmolVLA has been reported to match or exceed $\pi_0$ \citep{smolvla2025}, so rankings established there need not transfer here.}

Across policies, the failure mode distributions reveal a consistent pattern: \textbf{depth errors are the dominant failure mode}, accounting for 20--35\% of all test episodes, while lateral errors are markedly less frequent (0--25\%) and task breakdowns are essentially absent (0\%) in the baseline regime. Even the strongest model, $\pi_0$, fails almost exclusively due to depth errors (20\%) rather than lateral mislocalization (5\%). This pattern indicates that, under the dual-camera setup with full training data, the policies have largely solved the lateral localization problem but continue to struggle with the harder challenge of estimating the precise approach distance to a thin, low-contrast thread in 3D space.

\begin{figure}[!t]
  \centering
  \includegraphics[width=0.9\textwidth]{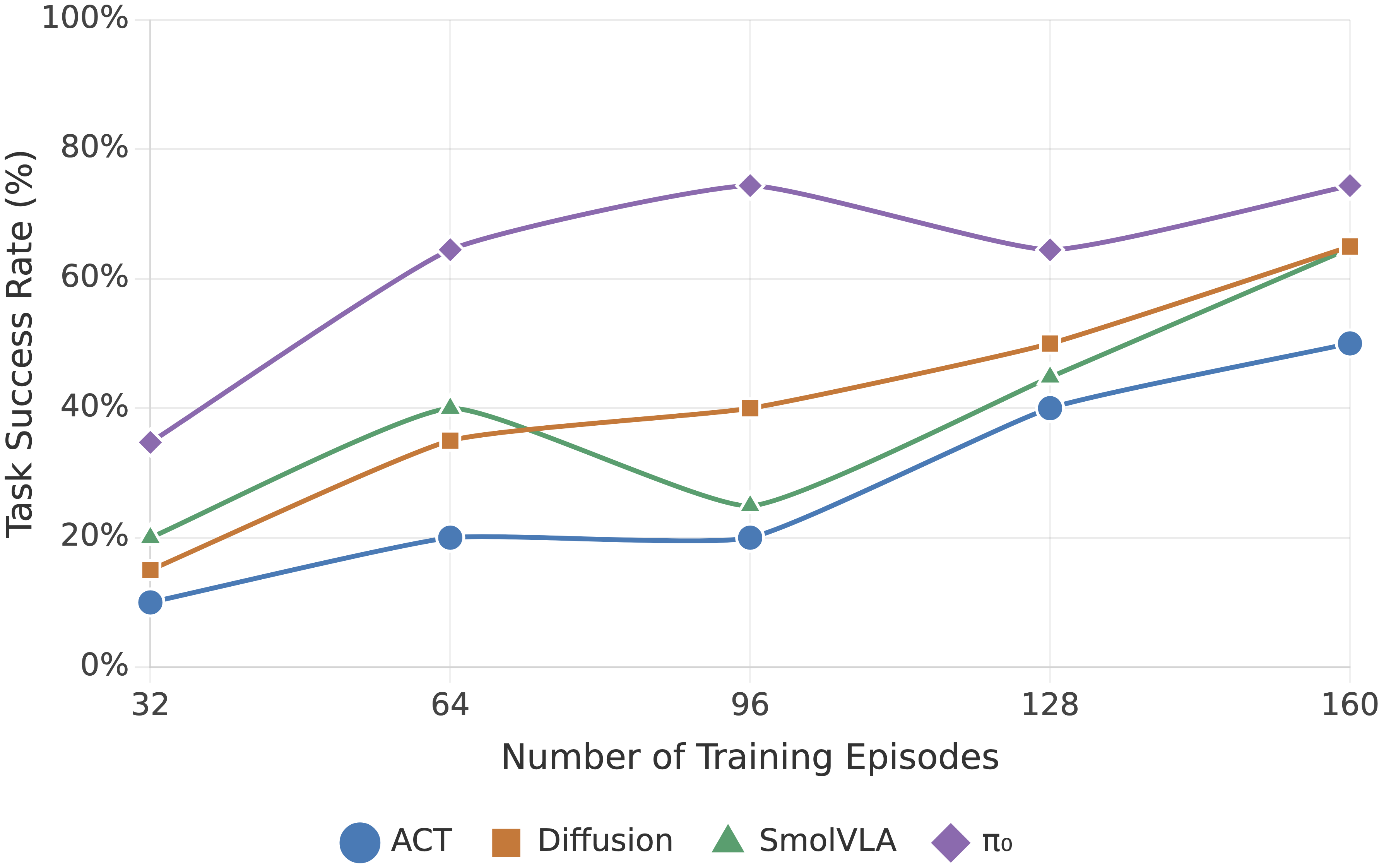}
  \caption{Task success rate as a function of training dataset size. $\pi_0$ achieves strong performance even at 96 episodes, while ACT and Diffusion Policy require substantially more data to reach competitive success rates.}
  \label{fig:data_scaling}
\end{figure}

\subsection{Effect of Dataset Size}
Figure~\ref{fig:data_scaling} shows the task success rate as a function of training dataset size for each policy. More demonstration data generally reduces task failure. ACT, Diffusion Policy, and SmolVLA generally improve from 32 to 160 episodes, though minor non-monotonic dips appear at individual data points (e.g., SmolVLA at 96 episodes, $\pi_0$ at 128 episodes), which we attribute to sampling variance over 20 test trials rather than a meaningful performance decrease. $\pi_0$, by contrast, already reaches 65\% success at just 64 episodes and stays in a narrow 65--75\% band for the rest of the curve, indicating that its pretrained backbone adapts to small amounts of in-domain data far more efficiently than the trained-from-scratch policies. The failure mode stacks in Figure~\ref{fig:failure_modes} reveal two clear trends: first, more data consistently reduces task breakdowns across all policies, and this effect is especially pronounced for Diffusion Policy, which suffers from a high breakdown rate at low data (25--30\% of episodes at 32--64 episodes) that fades only above $\sim$128 episodes. Second, depth errors remain the most stubborn failure category regardless of dataset size, accounting for 20--35\% of test episodes even at 160 episodes for every policy, suggesting that depth reasoning for thin, low-contrast objects is bottlenecked by the available perceptual signal rather than by demonstration count.

\begin{table}[t]
  \centering
  \small
  \caption{Camera view ablation (100\% training data, clean background). Failure modes as \% of all test episodes. Bold indicates dominant failure mode shift.}
  \begin{tabular*}{\textwidth}{@{\extracolsep{\fill}}llcccc}
  \toprule
    \textbf{Policy} & \textbf{Camera} & \textbf{Success (\%)} & \textbf{Lateral (\%)} & \textbf{Depth (\%)} & \textbf{Breakdown (\%)} \\
    \midrule
    \multirow{3}{*}{ACT}
      & Dual     & 50 & 25 & 25 &  0 \\
      & On-arm   & 20 & 10 & \textbf{65} &  5 \\
      & Side     & 40 & \textbf{35} & 20 &  5 \\
    \midrule
    \multirow{3}{*}{Diffusion}
      & Dual     & 65 &  0 & 35 &  0 \\
      & On-arm   & 40 &  0 & \textbf{50} & 10 \\
      & Side     & 20 & \textbf{35} & 20 & 25 \\
    \midrule
    \multirow{3}{*}{SmolVLA}
      & Dual     & 65 & 10 & 25 &  0 \\
      & On-arm   & 40 &  5 & \textbf{45} & 10 \\
      & Side     & 10 & \textbf{45} & 30 & 15 \\
    \midrule
    \multirow{3}{*}{$\pi_0$}
      & Dual     & 75 &  5 & 20 &  0 \\
      & On-arm   & 45 & 10 & \textbf{40} &  5 \\
      & Side     & 15 & \textbf{30} & 20 & \textbf{35} \\
    \bottomrule
  \end{tabular*}
  \label{tab:camera_ablation}
\end{table}

\begin{figure}[!htbp]
  \centering
  \includegraphics[width=\textwidth]{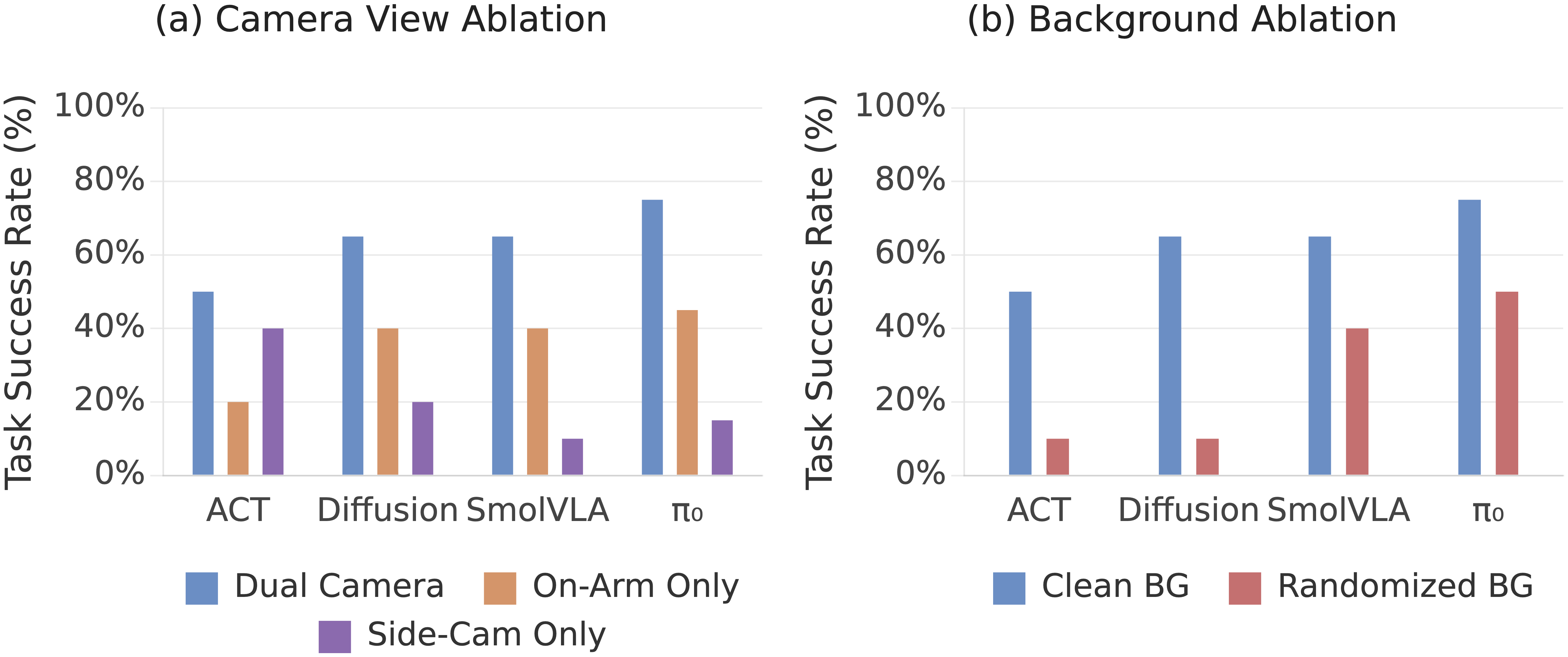}
  \caption{(a) Camera view ablation: all policies degrade under single-camera setups, but the pattern differs: on-arm only retains moderate performance for Diffusion and SmolVLA, while side-cam only is particularly detrimental for SmolVLA and $\pi_0$. (b) Background ablation: ACT and Diffusion Policy suffer severe degradation ($-$40 and $-$55 pp), while the pretrained VLA-based policies retain substantially higher performance.}
  \label{fig:ablation}
\end{figure}

\subsection{Effect of Camera Viewpoint}

Table~\ref{tab:camera_ablation} reports task success rates and failure mode distributions across camera configurations, while Figure~\ref{fig:ablation}(a) visualizes the success rate comparison. 
Removing either camera degrades success substantially for all four policies, confirming that the dual-camera setup is necessary. The two views are complementary: the side camera mainly supports depth estimation along the approach axis, while the on-arm camera mainly supports fine lateral alignment. The failure mode shifts in Table~\ref{tab:camera_ablation} reflect this cleanly. On-arm only (no side camera) causes depth errors to spike for every policy (40--65\%), while side-cam only (no on-arm camera) shifts the failure mass into lateral errors of 30--45\%.

% The two single-camera conditions are also not symmetric: for Diffusion, SmolVLA, and $\pi_0$, losing the on-arm camera is the more catastrophic of the two, dropping success to 10--20\% versus 40--45\% under on-arm-only. This indicates that lateral precision, contributed by the wrist-mounted view, is the dominant bottleneck for these policies, and supplies a lateral cue that the side camera cannot recover. ACT is the exception: it degrades roughly symmetrically and is in fact harmed more by losing the side camera, consistent with its smaller capacity and reliance on coarse spatial layout rather than fine wrist alignment.

\begin{table}[t]
  \centering
  \small
  \caption{Background ablation (dual camera, 100\% training data). Failure modes as \% of all test episodes. Bold indicates most prominent result per policy.}
  \begin{tabular*}{\textwidth}{@{\extracolsep{\fill}}llcccc}
  \toprule
    \textbf{Policy} & \textbf{Background} & \textbf{Success (\%)} & \textbf{Lateral (\%)} & \textbf{Depth (\%)} & \textbf{Breakdown (\%)} \\
    \midrule
    \multirow{2}{*}{ACT}
      & Clean      & 50 & 25 & 25 &  0 \\
      & Randomized & 10 & 15 & \textbf{70} &  5 \\
    \midrule
    \multirow{2}{*}{Diffusion}
      & Clean      & 65 &  0 & 35 &  0 \\
      & Randomized & 10 &  0 &  0 & \textbf{90} \\
    \midrule
    \multirow{2}{*}{SmolVLA}
      & Clean      & 65 & 10 & 25 &  0 \\
      & Randomized & 40 &  5 & \textbf{55} &  0 \\
    \midrule
    \multirow{2}{*}{$\pi_0$}
      & Clean      & 75 &  5 & 20 &  0 \\
      & Randomized & 50 & 10 & \textbf{40} &  0 \\
    \bottomrule
  \end{tabular*}
  \label{tab:env_robustness}
\end{table}

\subsection{Environmental Robustness}

Table~\ref{tab:env_robustness} and Figure~\ref{fig:ablation}b present the effect of background randomization. The table reports exact success rates and failure mode shifts; the figure visualizes the magnitude of the performance drop per policy.
Background randomization cleanly separates the four policies. The two VLA-based policies remain functional: $\pi_0$ retains 50\% success ($-$25\,pp from its 75\% baseline) and SmolVLA retains 40\% ($-$25\,pp), with their failures dominated by depth errors (40\% and 55\% respectively) rather than wholesale collapse. ACT and Diffusion Policy, by contrast, show clear overfitting to the training environment: ACT drops from 50\% to 10\% with depth errors spiking to 70\% (suggesting its encoder had latched onto background cues for depth that no longer hold), and Diffusion Policy collapses even more dramatically, from 65\% to 10\% with 90\% of episodes ending in task breakdown (no coherent attempt at all). These results indicate that frozen pretrained vision-language backbones provide visual representations that are more robust to background variation, allowing the VLA-based policies to degrade gracefully where the trained-from-scratch policies do not.

\subsection{Task Completion Time}

Table~\ref{tab:completion_time} reports the task completion time per episode under the baseline condition, with all policies evaluated under the same inference setup (Appendix~\ref{sec:inference}). These times serve as a relative comparison across architectures; absolute latencies would vary with hardware and optimization. $\pi_0$ completes episodes in 6.6\,s on average with low variance (std 0.9\,s), producing smooth, decisive trajectories with minimal hesitation. SmolVLA is moderately slower (10.9 $\pm$ 2.0\,s) with occasional mid-trajectory corrections. ACT takes 13.8 $\pm$ 1.6\,s, showing consistent but slower motion. Diffusion Policy is the slowest at 27.8 $\pm$ 9.1\,s with high variance (range 17--58\,s), reflecting the iterative denoising process that introduces perceptible latency and produces frequent stop-and-adjust micro-motions rather than fluid reaches. For collaborative surgical assistance, where the robot must coordinate with the surgeon's ongoing workflow, this difference is practically significant: $\pi_0$'s fast, smooth execution is more compatible with the rhythm of live suturing than the hesitant trajectories of the other policies.
\begin{table}[!htbp]
  \centering
  \small
  \caption{Task completion time per episode (baseline condition, $n=20$ trials each).}
  \begin{tabular*}{\textwidth}{@{\extracolsep{\fill}}lcc}
  \toprule
    \textbf{Policy} & \textbf{Mean $\pm$ Std (s)} & \textbf{Range (s)} \\
    \midrule
    $\pi_0$          & 6.6 $\pm$ 0.9 & [5, 9]  \\
    SmolVLA          & 10.9 $\pm$ 2.0 & [9, 15] \\
    ACT              & 13.8 $\pm$ 1.6 & [12, 17] \\
    Diffusion Policy & 27.8 $\pm$ 9.1 & [17, 58] \\
    \bottomrule
  \end{tabular*}
  \label{tab:completion_time}
\end{table}

\subsection{In-Loop Surgeon--Robot Suturing Trial}
\label{sec:inloop}

To complement the controlled benchmark evaluation, we conducted a small in-loop trial in which the strongest policy ($\pi_0$, dual camera, full data) cooperated with a surgeon performing real running suturing on a tissue phantom. Each \emph{round} consisted of \textbf{five consecutive running stitches} on the same wound, and a round was scored as successful only if the robot correctly followed the suture for all five stitches in a row without forcing the surgeon to stop or manually re-present the thread. We ran a total of five \emph{rounds} (25 stitches): three rounds completed all five stitches successfully, and two rounds failed on the fifth stitch when the gripper approached significantly off target, yielding 23 of 25 (92\%) individual stitches completed successfully.

The conditions of this in-loop trial differ in one important way from the controlled evaluation. The controlled evaluation scored an episode as successful only when the gripper made millimeter-precise contact with the thread on a stationary phantom, with no surgeon intervention permitted. In the in-loop setting, the surgeon was free to make minor natural adjustments that a human assistant would also depend on, such as nudging the thread by a few millimeters, provided these adjustments did not slow down the workflow or break the rhythm of the closure. As a result, episodes that would have been counted as marginal failures under the controlled evaluation are clinically usable in practice, because the surgeon naturally compensates for sub-centimeter offsets when receiving a thread from any assistant. The two failed rounds, by contrast, involved approach errors large enough to require the surgeon to release the needle driver or forceps to manually correct the robot, which we consider a workflow-breaking failure.

\section{Discussion}
\label{sec:discussion}
%% ====================================================================

This study set out to test whether general-purpose imitation learning policies, developed for tabletop manipulation, can support a fundamentally different use case: intra-operative collaboration with a surgeon during open surgery. Our results provide affirmative evidence. Four policies, trained on only 160 demonstrations, reach 50--75\% success on a controlled suture-following benchmark, and the strongest ($\pi_0$) retains $\geq$50\% success under unseen visual backgrounds. More importantly, the same policy cooperates with a surgeon to complete 23 of 25 stitches in a live suturing trial, demonstrating that surgeon--robot collaboration during an active procedure is feasible with current imitation learning methods and open-source hardware.

\paragraph{Architectural insights for surgical collaboration.} The performance ordering across our experiments reflects each policy's architectural fit for this collaborative setting. ACT ($\sim$52\,M parameters, trained from scratch) has the lowest capacity and struggles most with fine approach precision, explaining its lower baseline. Diffusion Policy is prone to task breakdown under low data and novel backgrounds, likely because its iterative denoising structure can settle into task-irrelevant trajectories when conditioning signals are unfamiliar. Additionally, the multi-step denoising process produces noticeably less smooth arm movements with frequent hesitation and micro-corrections, which is a practical concern for real-time surgical collaboration where the robot must move decisively alongside the surgeon. The two VLA-based policies (SmolVLA and $\pi_0$), which keep pretrained vision-language backbones frozen and fine-tune only the action expert, show markedly better robustness: their failures under distribution shift remain geometrically interpretable (depth errors) rather than degenerating into incoherent behavior. This suggests that pretrained visual representations, which encode scene understanding from large-scale data, are a particularly valuable foundation for policies that must operate in the visually variable conditions of a real surgical workflow. \response{We attribute this robustness to the VLA-based policies as configured, in which a pretrained vision-language backbone is an integral component, rather than to the vision-language backbone in isolation.}

\paragraph{From controlled evaluation to live collaboration.} The in-loop surgeon--robot trial (Section~\ref{sec:inloop}) provides initial evidence that the collaborative paradigm we study can translate beyond the lab. Throughout the trial, the surgeon operated with both hands occupied by the needle driver and forceps, cooperating with the robot as they would with a human assistant and without any additional personnel. Under this setup, we observe that the in-loop workflow naturally tolerates approach errors that the strict lab protocol penalizes, because surgeons adjust to sub-centimeter offsets from any assistant. This suggests that the practical usability of these policies in a collaborative setting is higher than the controlled benchmark numbers alone would indicate. The two failures (both on the fifth stitch of a round) involved errors large enough to interrupt the surgical workflow, pointing to gripper design and depth perception as the most impactful areas for improvement.

\paragraph{Clinical implication.} The central clinical contribution of this work is to establish that surgeon--robot collaborative assistance in open surgery is a viable research direction that can be studied with accessible hardware and existing ML methods. This matters because the global surgical workforce shortage \citep{meara2015global, kewalramani2025workforce} disproportionately affects the availability of trained assistants, particularly in community hospitals, surgical training programs, and low-resource settings. A robotic system that can reliably perform the assistant's repetitive subtasks during a live procedure would directly address this bottleneck. Even in well-staffed settings, consistent robotic assistance could reduce team size requirements, sustain quality across long procedures where human fatigue affects performance, and free the human assistant for tasks requiring higher clinical judgment. \response{We frame this as augmentation rather than replacement: lower-complexity tasks such as suture following also serve an educational role through which surgical assistants train, so the intended benefit is to offload repetitive workload and to support teams in under-resourced settings, not to remove the supervised training pathway through which assistants develop expertise.}

\paragraph{Limitations and future work.}
\label{sec:limitations}
This study evaluates a single assistive subtask on silicone tissue phantoms. Extending to biological tissue, other assistive subtasks (tension maintenance, instrument handover, tissue retraction), and the full complexity of a live OR are natural next steps, as is a clinical study with multiple surgeons and IRBs. \response{Quantitatively assessing the gentle tensioning pull, in particular, requires realistic tissue mechanics that a silicone phantom does not reproduce, making biological tissue the appropriate setting for evaluating tension regulation. All demonstrations were also collected by a single teleoperator to keep motion style consistent; collecting multi-operator data to test generalization across demonstration styles is a valuable extension, especially for more complex assistive tasks.} Depth perception remains the dominant residual failure mode and would benefit from stereo or depth-sensing cameras. A gripper tailored to thin, deformable thread, with a wider compliant jaw and a larger capture envelope, would also likely convert many of the near-miss lateral and depth failures we observed into successful grasps, since our current parallel gripper leaves little tolerance around the thread. The current hardware operates outside a sterile field; clinical use would require sterile draping and biocompatible surfaces, which are well-precedented in existing OR equipment. Prospective studies of surgeon--robot cooperation across longer procedures and varied tasks are needed to understand how trust and workflow adaptation develop in this collaborative paradigm.

%% ====================================================================
\section{Conclusion}
\label{sec:conclusion}
%% ====================================================================

We presented the first systematic evaluation of general-purpose imitation learning for surgeon--robot collaborative assistance in open surgery, using suture following as a case study. Our results suggest that imitation learning policies can be adapted to intra-operative surgical collaboration using accessible, non-specialized open-source hardware and a modest number of demonstrations, and a live surgeon--robot trial offers initial evidence for the practical viability of this approach. Pretrained vision-language backbones emerge as a particularly valuable architectural ingredient, and we identify depth perception and end-effector design as promising directions for future improvement. Taken together, these findings indicate that collaborative robotic assistance during open surgical procedures is a feasible and clinically meaningful target for imitation learning.

\acks{This research was supported in part by Nebius AI for Accelerating Research.}

%Do NOT change font size of references or modify the bibliography style
\bibliography{references}

\newpage
\appendix

\section*{Appendix A. Training Details}

All models are trained using the LeRobot default training recipe for each architecture, adapting only batch size to fit the available GPU memory. ACT, Diffusion Policy, and SmolVLA are trained on a single NVIDIA L40S GPU (46\,GB VRAM); $\pi_0$ is trained on a single NVIDIA H200 GPU (141\,GB VRAM). \response{All architecture-specific hyperparameters follow the LeRobot defaults for each policy.} Table~\ref{tab:hyperparams} summarizes the training hyperparameters and Table~\ref{tab:training_loss} reports the final training loss for all 28 model configurations. Note that lower loss does not necessarily correspond to higher task success, as overfitting to the training distribution can produce low loss but poor generalization.

\begin{table}[h]
  \centering
  \small
  \caption{Training hyperparameters and compute requirements.}
  \begin{tabular*}{\textwidth}{@{\extracolsep{\fill}}lcccc}
  \toprule
    & \textbf{ACT} & \textbf{Diffusion} & \textbf{$\pi_0$} & \textbf{SmolVLA} \\
    \midrule
    Visual encoder    & ResNet-18 & ResNet-18 & PaliGemma ViT (frozen) & SmolVLM ViT (frozen) \\
    Parameters (M)    & $\sim$52 & $\sim$263 & $\sim$3{,}500 & $\sim$450 \\
    Batch size        & 32 & 32 & 8 & 32 \\
    Training steps    & 30{,}000 & 70{,}000 & 40{,}000 & 50{,}000 \\
    Training time (h) & $\sim$6 & $\sim$9.3 & $\sim$4.9 & $\sim$8.7 \\
    GPU               & L40S 46\,GB & L40S 46\,GB & H200 141\,GB & L40S 46\,GB \\
    \bottomrule
  \end{tabular*}
  \label{tab:hyperparams}
\end{table}

\begin{table}[h]
  \centering
  \small
  \caption{Final training loss across all configurations (data fraction with dual camera, and camera ablations at 100\% data).}
  \begin{tabular*}{\textwidth}{@{\extracolsep{\fill}}lp{1.1cm}p{1.1cm}p{1.1cm}p{1.1cm}p{1.1cm}p{1.1cm}p{1.1cm}}
  \toprule
    \textbf{Model} & \centering\textbf{100\%} & \centering\textbf{20\%} & \centering\textbf{40\%} & \centering\textbf{60\%} & \centering\textbf{80\%} & \centering\textbf{On-arm} & \centering\arraybackslash\textbf{Side} \\
    \midrule
    ACT       & \centering 0.052 & \centering 0.036 & \centering 0.041 & \centering 0.045 & \centering 0.048 & \centering 0.058 & \centering\arraybackslash 0.051 \\
    Diffusion & \centering 0.007 & \centering 0.003 & \centering 0.005 & \centering 0.006 & \centering 0.007 & \centering 0.008 & \centering\arraybackslash 0.007 \\
    SmolVLA   & \centering 0.015 & \centering 0.006 & \centering 0.009 & \centering 0.011 & \centering 0.013 & \centering 0.016 & \centering\arraybackslash 0.015 \\
    $\pi_0$   & \centering 0.020 & \centering 0.009 & \centering 0.013 & \centering 0.016 & \centering 0.020 & \centering 0.028 & \centering\arraybackslash 0.044 \\
    \bottomrule
  \end{tabular*}
  \label{tab:training_loss}
\end{table}

\begin{figure}[h]
  \centering
  \includegraphics[width=0.48\textwidth]{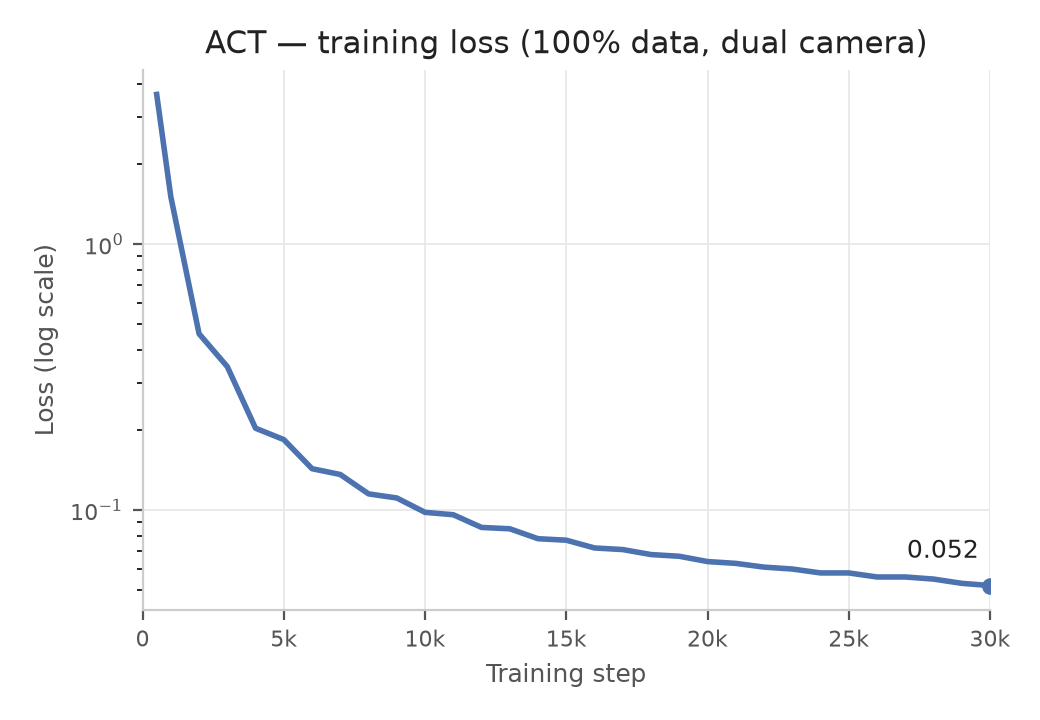}\hfill
  \includegraphics[width=0.48\textwidth]{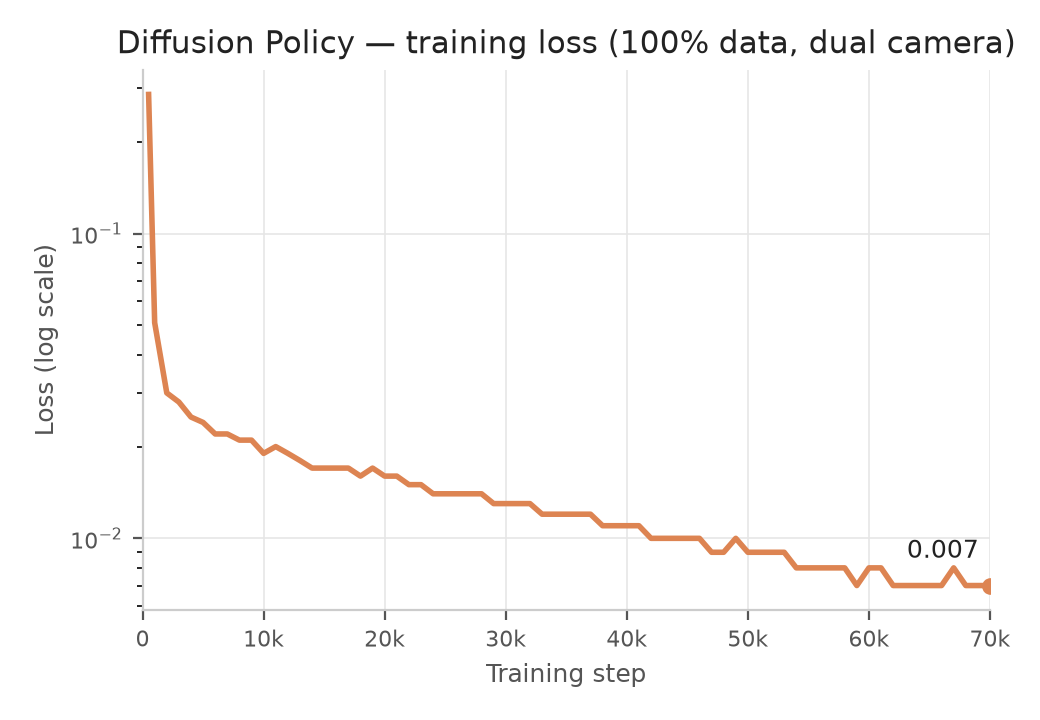}\\[4pt]
  \includegraphics[width=0.48\textwidth]{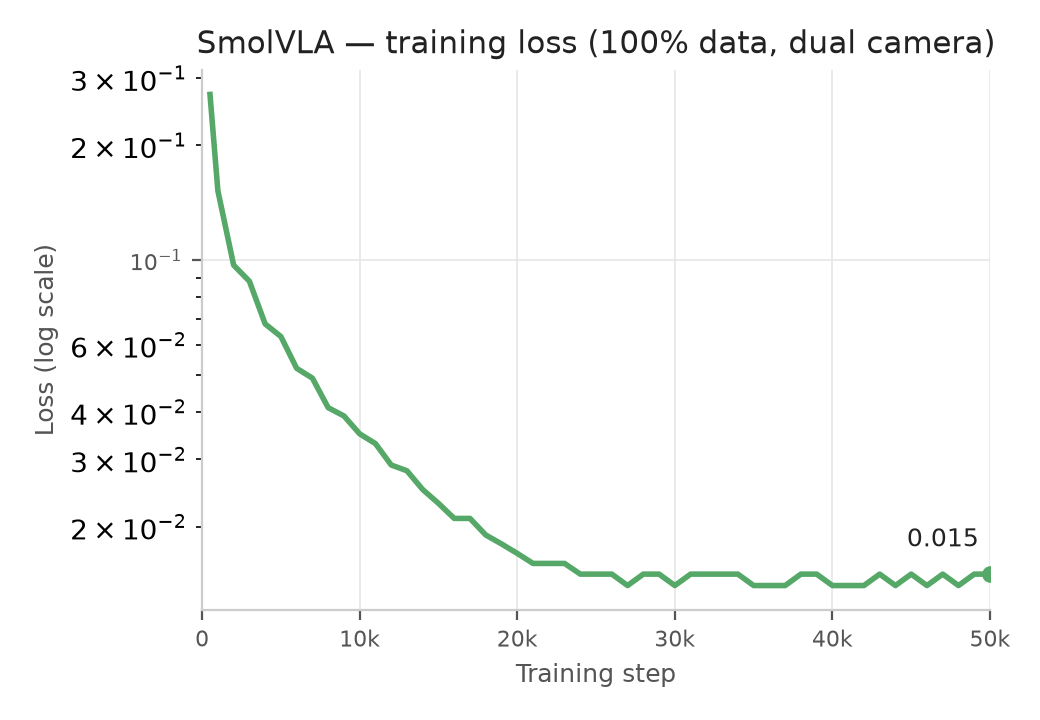}\hfill
  \includegraphics[width=0.48\textwidth]{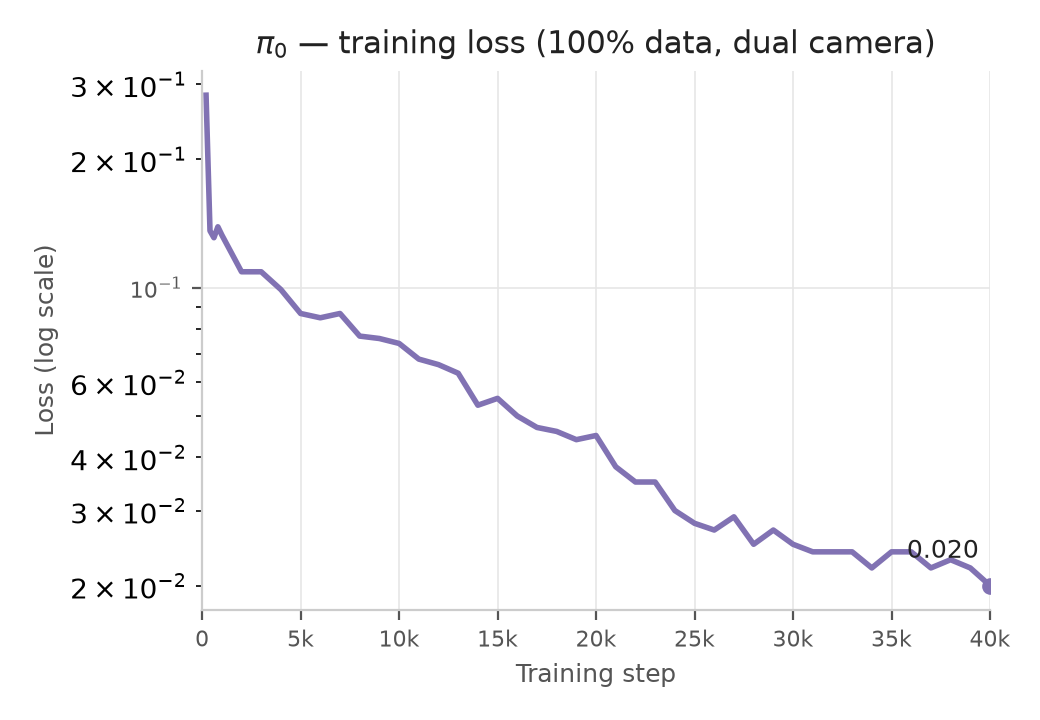}
  \caption{\response{Training loss versus optimization step for the four baseline policies (100\% data, dual camera). All losses decrease smoothly and plateau before the final step count, confirming convergence.}}
  \label{fig:training_curves}
\end{figure}

\response{Figure~\ref{fig:training_curves} shows the training loss curves for all models; the losses decrease smoothly and plateau well before the final step count, confirming that training has converged.}

\section*{Appendix B. Inference Details}
\label{sec:inference}

All policies are deployed on an Apple MacBook Pro (M1 Max, 64\,GB unified memory) with MPS (Metal Performance Shaders) acceleration under macOS 14.5, PyTorch 2.10.0, and LeRobot 0.5.1.

\end{document}